\documentclass[letterpaper, 10 pt, conference]{ieeeconf}  
\usepackage{graphicx}
\usepackage{cite}
\usepackage{array}
\usepackage{multirow}
\usepackage[table]{xcolor}
\usepackage{amsmath} 
\usepackage{caption}
\usepackage{threeparttable}
\usepackage{cuted}   
\usepackage{capt-of}  
\graphicspath{{Images/}{Images/ProfilePic/}{Images/IMUPaper/}}

\usepackage{booktabs,tabularx,pifont}
\newcommand{\cmark}{\ding{51}} 
\newcommand{\xmark}{\ding{55}}

\usepackage{tikz}

\IEEEoverridecommandlockouts                              

\makeatletter
\let\NAT@parse\undefined
\makeatother
\usepackage[hidelinks]{hyperref}

\title{\LARGE \bf
RoSHI: A Versatile Robot-oriented Suit for Human Data In-the-Wild 
}
\author{Wenjing Margaret Mao$^{*}$, Jefferson Ng$^{*}$, Luyang Hu$^{*}$, Daniel Gehrig, Antonio Loquercio\vspace{-4ex}%
\thanks{$^{*}$Equal contribution.}%
\thanks{Department of Electrical and Systems Engineering, University of Pennsylvania. 
Email: \{mwenjing, jefferzn, huluyang, dgehrig, aloque\}@seas.upenn.edu}%
}

\newcommand{\sysname}{RoSHI}
\begin{document}
\maketitle
\setlength\stripsep{0pt plus 0pt minus 0pt}

{\setlength{\abovecaptionskip}{2pt}
 \setlength{\belowcaptionskip}{0pt}
 \begin{strip}
   \centering
   \includegraphics[width=0.8\textwidth]{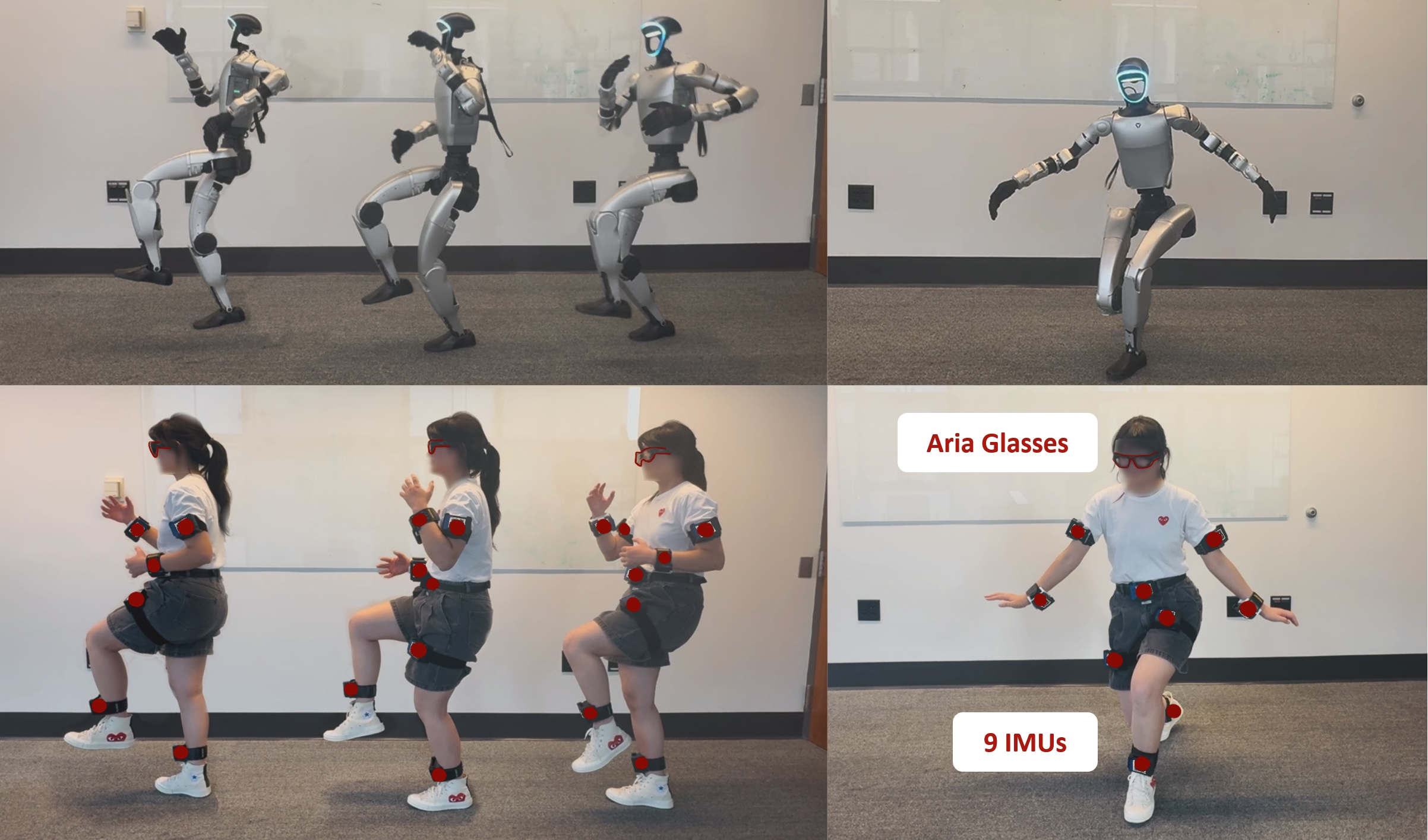}%
   \vspace{2ex}
   \captionof{figure}{Illustration of RoSHI, a \textbf{R}obot-\textbf{o}riented \textbf{S}uit for \textbf{H}uman Data \textbf{I}n-the-Wild. RoSHI is a low-cost, portable system for in-the-wild human motion capture (bottom row), and deployment of learned policies on a humanoid robot (top row). On the left, the robot executes alternating single-leg jumps; on the right, it performs a bowing motion. RoSHI fuses signals from Project Aria glasses and nine Inertial Measurement Units (IMUs) to capture synchronized 3D body pose, egocentric RGB video, and a globally consistent root trajectory. These multimodal signals emphasize reliability and long-horizon stability over per-frame accuracy, enabling dynamic human demonstrations to be successfully transferred into deployable humanoid skills. The third-person view is only for visualization and not used by our system.}
   \vspace{1em}
   \label{fig:profile-wide}
 \end{strip}
}


\thispagestyle{empty}
\pagestyle{empty}

\begin{abstract}
Scaling up robot learning will likely require human data containing rich and long-horizon interactions in the wild. Existing approaches for collecting such data trade off portability, robustness to occlusion, and global consistency. We introduce \emph{\sysname}, a hybrid wearable that fuses low-cost sparse IMUs with the Project Aria glasses to estimate the full 3D pose and body shape of the wearer in a metric global coordinate frame from \emph{egocentric} perception. This system is motivated by the complementarity of the two sensors: IMUs provide robustness to occlusions and high-speed motions, while egocentric SLAM anchors long-horizon motion and stabilizes upper body pose. We collect a dataset of agile activities to evaluate RoSHI. On this dataset, we generally outperform other egocentric baselines and perform comparably to a state-of-the-art exocentric baseline (SAM3D). Finally, we demonstrate that the motion data recorded from our system are suitable for real-world humanoid policy learning. For videos, data and more, visit the project webpage: \href{https://roshi-mocap.github.io/}{https://roshi-mocap.github.io/}


\end{abstract}


\section{INTRODUCTION}
\setlength{\parindent}{1.5em} 
\setlength{\parskip}{0pt}     








Robot learning increasingly depends on human data that can be collected easily, outside controlled labs, and at low cost. Existing options to collect such data, however, each involve trade-offs (Table~\ref{tab:capture_comparison}): marker-based systems (e.g., Vicon) require instrumented spaces and are expensive; commercial IMU suits (e.g., Xsens~\cite{roetenberg2009xsens}) are costly and often lack global localization; and vision-only pipelines~\cite{goel2023humans,pavlakos2023reconstructinghands3dtransformers,li2020simple} depend heavily on camera placement and scene conditions, with occlusions and lighting forcing practitioners to carefully curate clips and environments. As a result, there is still no widely adopted, humanoid-focused, portable full-body capture tool: an analogue to low-friction data interfaces like UMI~\cite{chi2024universalmanipulationinterfaceinthewild} or the Stick~\cite{shafiullah2023bringing} in manipulation.

Concretely, we study how to collect metric 3D human data in uninstrumented everyday environments for humanoid policy learning.
Building on the requirements of prior work in human-to-robot transfer~\cite{fu2024humanplushumanoidshadowingimitation,Peng_2018,he2025asapaligningsimulationrealworld,kareer2024egomimicscalingimitationlearning,zhu2025emma}, our desiderata comprise three synchronized signals: \textit{(i)} 3D human body pose~\cite{SMPL:2015}; \textit{(ii)} a globally consistent 6-DoF root trajectory, and \textit{(iii)} egocentric RGB video.

To obtain these signals, we propose RoSHI, a \textbf{R}obot-\textbf{o}riented \textbf{S}uit for \textbf{H}uman Data \textbf{I}n-the-Wild. RoSHI is a hybrid wearable data-collection system that combines low-cost inertial trackers with egocentric sensing from Project Aria glasses~\cite{engel2023projectarianewtool}.
The IMUs estimate the wearer’s posture, providing robustness to visual occlusion, while the glasses supply two complementary streams: egocentric SLAM for global root trajectory and RGB video, which supports policy learning. 

\begin{figure*}[t]
  \centering
  \includegraphics[width=\textwidth]{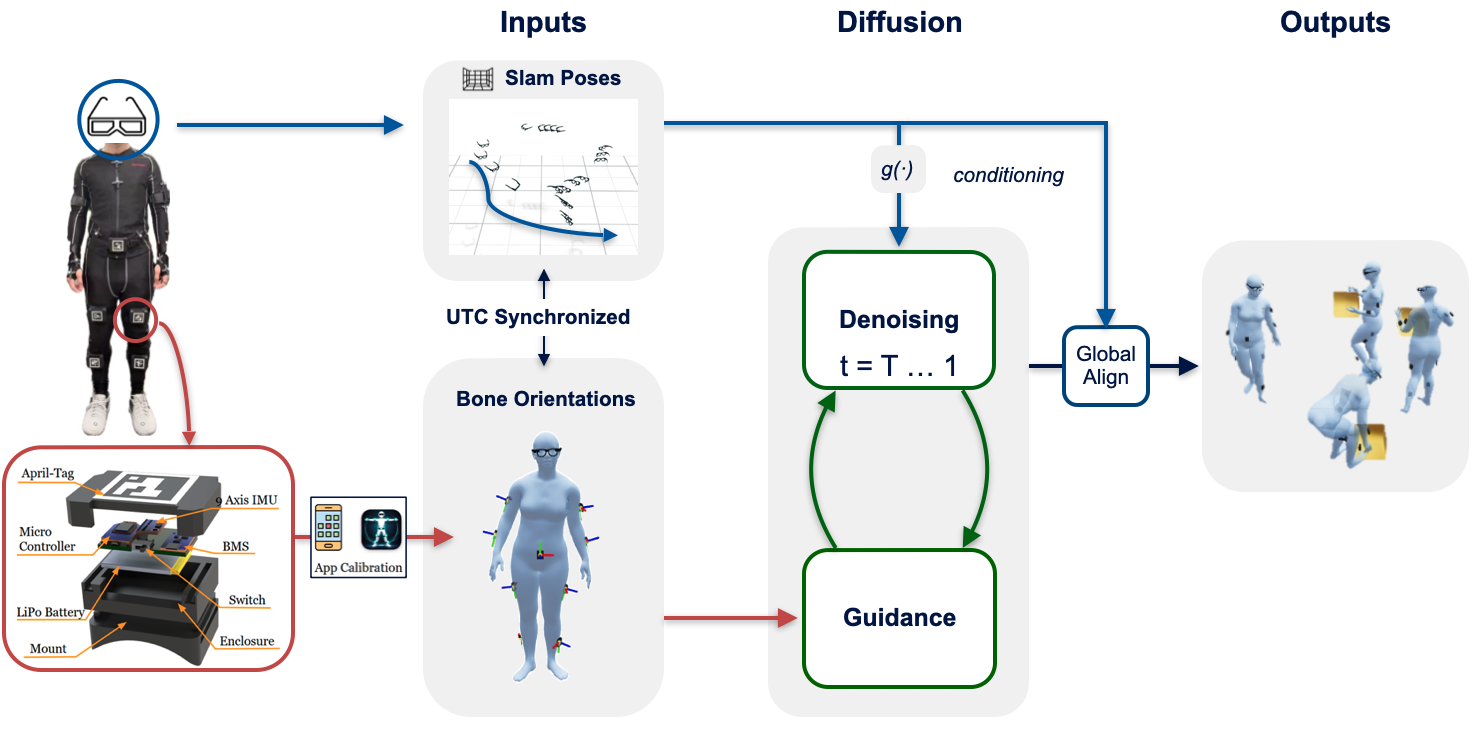}
  \caption{Overview of the RoSHI data pipeline. A user wears a low-cost, portable suit comprising nine IMU trackers (bottom left), and a Project Aria headset. Each tracker  integrates a microcontroller, 9-axis IMU, battery management system, and LiPo battery within a compact enclosure. At a fixed, known offset, AprilTags are mounted on the enclosure to facilitate calibration. After an initial calibration procedure using an external iPhone, and custom APP, each IMU tracker provides bone orientations (two per arm and length, and pelvis) mapped to the pelvis world frame. Using the Project Aria SLAM poses as conditioning and IMU tracker outputs as guidance, we use the diffusion model in~\cite{yi2025egoallo} to generate articulated human poses, in the local Aria camera frame. We finally use the SLAM poses to map these poses to a global frame.}
  \label{fig:ph1}
\end{figure*}

The combined system, whose components are shown in Fig.~\ref{fig:ph1}, is portable, robot-agnostic, and enables in-the-wild capture without cages, external cameras, or pre-instrumented spaces. Importantly, the IMU subsystem relies on off-the-shelf consumer-grade inertial sensors rather than high-precision commercial motion-capture units, reducing the total hardware cost to approximately \$350 USD for nine IMUs and a USB receiver. It is also easily extensible: additional signals from the Aria glasses (e.g., eye gaze or audio) or entirely new sensors (e.g., depth or tactile sensors) can be incorporated with minimal effort.
Moreover, the system is modular: the IMU modules can be upgraded or replaced with alternative off-the-shelf designs, such as those described in \cite{slimevr_tracker_esp_v054} or other open-source implementations. Similarly, one can replace the Aria glasses with alternative state-estimation cameras~\cite{zabatani2019intel,abdelsalam2024depth} or even standard RGB cameras paired with open-source SLAM algorithms.
%



We design a system that is cheap, portable, and capable of operating over long horizons. This features make our system compatible with large scale data collection efforts for robotics applications while requiring minimal infrastructure and cost. Despite its minimalist design the system still produces reliable, high-quality, and physically plausible human motion data, which can be smoothly transferred from human motion to humanoid robots.



To achieve high-quality data collection while not compromising reliability and long-range operation, we
leverage complementary visual and inertial sensing. Prior vision-based methods provide accurate 6 DoF global trajectories and 3D articulated pose estimates of the wearer's body, but fail when the subject is occluded for long periods of time, or blurred during fast motion~\cite{goel2023humans,rajasegaran2022tracking,pavlakos2023reconstructinghands3dtransformers}. IMUs remain robust under occlusion but suffer from drift without external anchors. \sysname\ mitigates both failure modes by combining these modalities.

We show that, quantitatively, RoSHI has a lower mean per-joint position error than state-of-the-art IMU-only and egocentric baselines across a diverse set of motions, demonstrating consistent improvements in both global joint localization and articulated pose reconstruction. We also show qualitatively that data collected with our system is suitable for policy training and deployment on a humanoid robot, as shown in Fig.~\ref{fig:profile-wide}.



\begin{table}[t]
\small
\centering

\setlength{\tabcolsep}{2pt}
\renewcommand{\arraystretch}{1}
\begin{tabularx}{\columnwidth}{@{} l
  >{\centering\arraybackslash}X
  >{\centering\arraybackslash}X
  >{\centering\arraybackslash}X
  >{\centering\arraybackslash}X
  >{\centering\arraybackslash}X @{}}
\toprule
\textbf{Type} & \textbf{View} & \textbf{Occlusion Robust} & \textbf{Global traj.} & \textbf{Portability} & \textbf{Cost} \\
\midrule
Vicon-based           & 3rd      & \cmark & \cmark & Low   & High\\
Inertial suit  &  Wear & \cmark & \xmark & High & Varies \\
Video      & 3rd      & \xmark & \cmark & High  & Low  \\
Egocentric  & 1st      & \xmark & \cmark & High & Low \\
\textbf{Ours (\sysname)}  & 1st+Wear & \cmark & \cmark & High   & Low \\
\bottomrule
\end{tabularx}
\caption{Comparison of Collection Systems for Whole-Body Data. \label{tab:capture_comparison}}
\vspace{-3ex}
\end{table}

\noindent\textbf{Contributions.}
\textit{(i)} A portable, low-cost, robot-oriented capture pipeline that fuses sparse IMUs with egocentric sensing, enabling in-the-wild collection of synchronized 3D human hand and body pose, metric global trajectory, and first-person RGB video;
\textit{(ii)} A lightweight human pose generation approach conditioned on SLAM poses and guided by bone orientations, remaining robust to vision occlusions;
\textit{(iii)} An open-source release of our hardware design and full data-collection stack, together with a curated dataset of diverse whole-body motions with annotated ground truth.




\section{RELATED WORK}
\setlength{\parindent}{1.5em} 
\setlength{\parskip}{0pt}     


Third-person vision is a standard way to reconstruct a 3D human pose. Such methods (single- or multi-view) use temporal priors and learned regressors to achieve strong per-frame quality~\cite{kocabas2021pareattentionregressor3d, sun2021monocularonestageregressionmultiple, kolotouros2019learningreconstruct3dhuman, li2021deeptwostreamvideoinference,li2020simple,pavlakos2023reconstructinghands3dtransformers,goel2023humans,rajasegaran2022tracking}. SAM 3D Body, for example, recovers metric-scale full-body meshes in a calibrated third-person setup and serves as a strong vision-based baseline in our evaluation~\cite{yang2026sam3dbody}. These methods can bridge short occlusions but are sensitive to distance between the camera and the human, lighting, and/or clutter.
Methods targeting dynamic cameras and occlusion help, but do not resolve long-horizon ambiguity in the absence of anchors~\cite{yuan2022glamrglobalocclusionawarehuman}. Although highly accurate in controlled environments, third-person pipelines are constrained by camera placement and field-of-view, making them not truly portable and, therefore, ill-suited as a primary interface for a unified in-the-wild data collection tool.

In contrast, egocentric sensing opens several interesting opportunities for data collection. Existing large first-person datasets provide RGB and activity labels at scale~\cite{grauman2022ego4dworld3000hours, Damen2018EPICKITCHENS, damen2020epickitchensdatasetcollectionchallenges}, with emerging efforts toward collecting richer daily motion in the wild~\cite{ma2024nymeriamassivecollectionmultimodal, yang2025egolifeegocentriclifeassistant}. Recent AR headsets (e.g., Project Aria) add headset-grade SLAM, scene graphs, and wrist/palm keypoints~\cite{engel2023projectarianewtool}, enabling downstream body pose and hand–object interaction estimation~\cite{kareer2024egomimicscalingimitationlearning}. EgoAllo further extends this direction by directly estimating the wearer's full-body pose and mesh from Aria Glasses, making it a feasible foundation for portable motion data collection~\cite{yi2025egoallo}. However, as we show in our experiments, it struggles when large parts of the body are not visible from egocentric vision, or during high-speed motions.

Exocentric vision systems are the standard approach for obtaining ground-truth motion capture. Marker-based studio systems such as Vicon and OptiTrack use multiple calibrated externally mounted infrared cameras to track reflective markers and triangulate precise 3D joint trajectories within a fixed capture volume~\cite{optitrack_website,vicon_website}. With carefully placed markers in a motion-capture suit, these systems can produce a highly accurate body pose. However, they require instrumented spaces, careful calibration, and substantial cost, which limits scalability and makes in-the-wild collection impractical, relegating them primarily to use as ground truth for benchmarking other methodologies.

IMU motion capture estimates full-body pose by attaching a small number of inertial sensors to key body segments and reconstructing the remaining joints through a kinematic model (often aided by learned motion priors). Commercial suits such as Xsens~\cite{roetenberg2009xsens} and Noitom~\cite{noitom_website} follow this paradigm but use higher-accuracy IMU hardware, yielding strong joint-orientation tracking for dynamic motions at substantial cost (e.g., roughly \$4,500 USD to \$14,000 USD for different Xsens configurations)~\cite{roetenberg2009xsens}. However, even these high-end systems typically lack true global localization as drift accumulates over long horizons~\cite{Ziegler2011AccurateHM}. At the lower-cost end, DIY and community IMU designs such as SlimeVR~\cite{slimevr_tracker_esp_v054} improve accessibility but rely on noisier consumer IMU chips and hence commonly require frequent user-driven calibration (e.g. T-pose initialization or periodic resets) to manage heading drift. In addition to that, while high-end suits can often rely on relatively accurate inertial signals to produce usable motion tracking despite long-horizon drift, low-cost IMU set ups only support local body pose tracking and do not yield reliable metric-scale global trajectories, requiring external anchors or complementary sensing to recover globally consistent motion.

The goal of our work is combining the advantages of egocentric inertial and vision systems. We do so by developing a fusion framework to integrate the signals from sparse consumer-grade IMUs with egocentric position and body pose estimation to achieve globally consistent, long-horizon human motion
capture.

\section{Method}
\setlength{\parindent}{1.5em} 
\setlength{\parskip}{0pt}     

\subsection{Overview}
In the following, we provide details of the components of our system (see Fig.~\ref{fig:ph1}).
It comprises nine low-cost IMU trackers inspired by open-source designs such as SlimeVR~\cite{slimevr_tracker_esp_v054}, and Project Aria glasses~\cite{engel2023projectarianewtool} providing a wide-angle egocentric RGB video stream. Together this yields a portable solution for outdoor motion capture.  We leverage data collected from these components to estimate \textit{(i)} the 6-DoF headset trajectory ${}^CT_{W_c}(t)=({}^C R_{W_c}(t),{}^C p_{W_c}(t))$ (directly provided by the glasses), and \textit{(ii)} 3D body poses (SMPL~\cite{SMPL:2015}), all expressed in global frame $W_c$ defined at the start of headset recording. To derive body poses, we first leverage the EgoAllo diffusion model~\cite{yi2025egoallo} which generates them in the camera frame $C$, and we then transform them into the global frame $W_c$ via ${}^CT_{W_c}(t)$. Natively, ~\cite{yi2025egoallo} conditions diffusion on 6 DoF trajectory ${}^CT_{W_c}(t)$ and guides it with hand poses estimated from the video stream via HaMeR~\cite{pavlakos2023reconstructinghands3dtransformers}. RoSHI simplifies this by instead guiding diffusion via bone orientations ${}^{W_p} R_{B_i}(t)$ with $i=1,...,9$ (two per arm, two per leg, and pelvis), derived from the IMU trackers. Here $W_p$ denotes the gravity-aligned frame of the pelvis IMU tracker. We will discuss these trackers, and their role in body pose generation next. 

\subsection{Hardware Components and Synchronization}
Each IMU tracker is assembled from standard breakout boards, integrating a 2.4\,GHz communication-enabled microcontroller, a battery management system (BMS), and a 9-axis IMU. We select the BNO085, a \$20 USD IMU chip with onboard sensor fusion, outputting orientations ${}^{W_i}R_{S_i}(t)$ without relying on external fusion algorithms. Here $S_i$ and $W_i$ are the sensor frame, and gravity-aligned local world frame of the $i^\text{th}$ IMU.
Using inexpensive components, the cost of each tracker is approximately \$30 USD, with up to 10 hours of continuous operation. A lightweight 3D-printed enclosure secures the PCB to reduce vibration during dynamic motion.

For real-world deployment, data from each tracker are transmitted to a custom USB dongle via a low-latency 2.4\,GHz peer-to-peer link, supporting communication ranges up to 100 meters in open space. The receiver aggregates orientation data at 100\,Hz. While our current configuration employs nine IMUs the system can be scaled depending on task requirements.
Instead of integrating the SlimeVR firmware stack, we directly transmit rotation and battery data to the workstation, enabling time-synchronized recording within our robotics data pipeline. This design reflects our hypothesis that low-cost IMU signals are sufficient for robot-ready data collection.

\noindent\textbf{Synchronization:} Reliable multimodal fusion requires accurate alignment across sensor streams. We synchronize three components: \textit{(i)} RGB video, \textit{(ii)} IMU measurements, and \textit{(iii)} Aria metadata. We timestamp all streams with a Unix time (UTC) clock and align them by nearest-neighbor matching in time. This is robust to different sampling rates: the video/SLAM streams run at $\sim$30\,Hz, whereas the IMU runs at 100\,Hz. It is also robust to occasional frame-rate drops when the app or Aria runs slower than the nominal setting. Since all devices are connected to the same LAN, the inter-device clock offset is typically below 100\,ms, which we treat as negligible at human-perceptual timescales.

\subsection{Body Pose Generation}
RoSHI generates body poses using the diffusion model in ~\cite{yi2025egoallo}, trained on AMASS~\cite{mahmood2019amass} with synthesized headset trajectories. During inference it guides diffusion to remain consistent with priors such as ground contact~\cite{yi2025egoallo,ci2023gfpose,zhang2023probabilistic}, as well as three complementary constraints based on estimated bone orientations ${}^{W_p} R_{B_i}(t)$ (see Fig.~\ref{fig:ph1}). First, the joint angles of the elbow, hip, and knee orientations are directly observable and thus compared to the diffusion model’s predictions. Second, for joints that are not directly observable, we exploit relative orientations. In particular, we compare the relative rotation between the pelvis and shoulder,  to the corresponding relative rotation implied by the diffusion model’s predictions via the body’s kinematic tree. Third, we enforce consistency between the relative pelvis-joint rotation across consecutive frames and the relative rotation predicted by the model. While similar constraints could be applied to other joints, we empirically found them unhelpful and omitted them for simplicity. Note that, since we only guide based on relative bone orientations, the pelvis world frame $W_p$ is arbitrary. Next we discuss how to estimate ${}^{W_p} R_{B_i}(t)$.

\noindent\textbf{Bone Orientation Tracking:} 
We estimate bone orientations by combining three transforms 
\begin{equation}
\label{eq:tracking}
    {}^{W_p}\!R_{B_i}(t) = {}^{W_p}\!R_{W_i}\;{}^{W_i}\!R_{S_i}(t)\;\left({}^{B_i}\!R_{S_i}\right)^{\top}.
\end{equation}
Here ${}^{W_i}\!R_{S_i}(t)$ comes directly from the IMU, and ${}^{W_p}\!R_{W_i}$ and ${}^{B_i}\!R_{S_i}$ are derived from calibration. Calibration is one of the main pain points in IMU-based motion capture. Standard methods rely on two-step calibration: First, a box calibration procedure  aligns all sensors to a shared reference, yielding ${}^{W_p}\!R_{W_i}$). Second, on-body registration via prescribed poses (typically T-pose/A-pose) is used to estimate the fixed offset ${}^{B_i}\!R_{S_i}$ between each IMU and its corresponding SMPL bone. In practice, pose-based calibration is brittle: an ideal T-pose is hard to reproduce consistently due to body shape, clothing, or uneven ground, so small pose deviations can introduce systematic offsets that persist throughout the session. Box calibration also makes recalibration inconvenient, since users must remove the IMUs, place them into the box, and re-wear them, discouraging recalibration even though strap slippage and IMU heading drift make it desirable. Next, we will show how RoSHI overcomes these challenges by leveraging an auxiliary video stream.

\subsection{Calibration}
RoSHI addresses these challenges with a vision-assisted calibration that can be performed while wearing the suit. We rigidly attach a 4\,cm AprilTag (with frame $T_i$) to each of the nine IMUs, with known rigid rotation ${}^{T_i}\!R_{S_i}$ between the tag and IMU sensor frame.
We record a short (20 to 40 second) video of natural motion using an iPhone (with camera frame $C_s$), running our custom app, while also logging IMU measurements ${}^{W_i}\!R_{S_i}(t)$. The app detects AprilTags using the AprilTag library~\cite{olson2011tags} and recovers the per-frame tag orientations ${}^{C_s}\!R_{T_i}(t)$. We run SAM 3D Body~\cite{yang2026sam3dbody} on the calibration video to estimate per-frame joint rotations, then convert its MHR outputs to the SMPL convention to obtain SMPL-aligned bone rotations ${}^{C_s}\!R_{B_i}(t)$ in the camera frame. We use this data to estimate ${}^{B_i}\!{R}_{S_i}$ and ${}^{W_p}\!{R}_{W_i}$ in Eq.~\eqref{eq:tracking}.

\noindent\textbf{Estimating ${}^{B_i}\!{R}_{S_i}$:}
For frames where both the tag detection is valid and the body pose estimate is confident, we compute per-frame point estimates
\begin{equation}
{}^{B_i}\!R_{S_i}(t) = \left({}^{C_s}\!R_{B_i}(t)\right)^{\top}\,{}^{C_s}\!R_{T_i}(t){}^{T_i}\!R_{S_i}.
\end{equation}
Since the bone-to-sensor rotation is assumed to stay constant over time, we filter these measurements by computing their Barycenter on $SO(3)$ via the Karcher mean
\begin{align}
{}^{B_i}\!\bar{R}_{S_i} &= \arg\min_{R\in SO(3)} \sum_{j=1}^{N_i} d_g\!\left(R,\;{}^{B_i}\!R_{S_i}(t_j)\right)^{2}\\
d_g(R_1,R_2)&=\left\lVert \log\!\left(R_1^{\top}R_2\right)\right\rVert .
\end{align}
where $N_i$ counts the valid detections of the $i^\text{th}$ tag and bone. 

\noindent\textbf{Estimating ${}^{W_p}\!{R}_{W_i}$:} 
To replace box calibration, we estimate a heading alignment that maps each IMU's local world frame $W_i$ into a shared reference world $W_p$ (defined as the local world of the pelvis). The tag expressed in $W_i$ is
\begin{equation}
{}^{W_i}\!R_{T_i}(t) = {}^{W_i}\!R_{S_i}(t)\;{}^{S_i}\!R_{T_i},
\end{equation}
so the world frame orientation in camera coordinates is
\begin{equation}
{}^{C_s}\!R_{W_i}(t) = {}^{C_s}\!R_{T_i}(t)\,\left({}^{W_i}\!R_{T_i}(t)\right)^{\top}.
\end{equation}
The Barycenter over frames where tag $i$ is visible yields ${}^{C_s}\!\bar{R}_{W_i}$ and, in particular for the pelvis, ${}^{C_s}\!\bar{R}_{W_p}$. Thus,
\begin{equation}
{}^{W_p}\!R_{W_i} = \left({}^{C_s}\!\bar{R}_{W_p}\right)^{\top}{}^{C_s}\!\bar{R}_{W_i}.
\end{equation}
These steps yield both on-body registration and cross-sensor world alignment without a calibration box or prescribed poses, enabling quick recalibration at any time without removing the IMUs.

\section{EXPERIMENTS}

To validate our system we collected human data across diverse activities spanning both indoor and outdoor settings (Sec.~\ref{sec:data}), and validated it's accuracy both qualitatively, and quantitatively against state-of-the-art (Sec.~\ref{sec:results}). Finally, we show that our lightweight hardware setup is practical, and provided sufficient accuracy for humanoid research (Sec.~\ref{sec:humanoid}).

Our evaluation seeks to answer three key questions: \textit{(i)} How do the body pose estimates produced by our system qualitatively compare against baseline methods? \textit{(ii)} Can sequences collected from RoSHI be effectively retargeted to a robot? \textit{(iii)} Can the collected data support the learning of whole-body control policies that successfully transfer to a physical humanoid?

\subsection{Experimental Setup}
\label{sec:data}
\textbf{Dataset:} We show the versatility of our system by collecting 11 motion sequences (Tab.~\ref{tab:dataset_activities}) performed by two data collectors. We organize these sequences into three datasets because they represent distinct motion regimes with different dominant failure modes for sensing and reconstruction. Dataset~1 contains primarily in-place motions with minimal occlusion, where SLAM position drift is less impactful and we can more directly evaluate body reconstruction quality and compare against strong vision-only baselines under favorable visibility. Dataset~2 contains motions with clear global translation and moderate agility, which increases reliance on global localization and can lead to intermittent field-of-view loss for vision-based baselines; correspondingly, we observe reduced vision-only reconstruction recall in this regime (SAM3D Rec., Tab.~\ref{tab:dataset_activities}). Dataset~3 contains the most agile, sports-like activities with fast direction changes and higher-speed dynamics; these sequences stress-test the IMU's ability to capture high-speed motion and the synchronization across modalities, while also challenging vision systems due to motion blur. 

\begin{table}[t]
\centering
\caption{Dataset composition and per-clip statistics. We report evaluation duration and SAM3D recall.}
\label{tab:dataset_activities}
\small
\setlength{\tabcolsep}{3pt}
\renewcommand{\arraystretch}{0.95}
\begin{tabular}{c l c c}
\toprule
\textbf{\#} & \textbf{Clip (activity)} & \textbf{Eval (s)} & \textbf{SAM3D Rec.} \\
\midrule
\multicolumn{4}{l}{\emph{Dataset 1}} \\
1  & \texttt{walk\_march\_jog\_run} & 36.0 & 100.0\% \\
2  & \texttt{stretch\_boxing\_bow\_wave} & 99.0 & 100.0\% \\
3  & \texttt{jumpjack\_squat\_oneleg} & 61.8 & 100.0\% \\
4  & \texttt{pick-up-box} & 49.0 & 99.9\% \\
\midrule
\multicolumn{4}{l}{\emph{Dataset 2}} \\
5  & \texttt{walk-sayhi-walk} & 74.3 & 77.0\% \\
6  & \texttt{pickup-walkaround} & 53.7 & 50.0\% \\
7  & \texttt{walk-jog-back-and-forth} & 41.1 & 100.0\% \\
8  & \texttt{jump-around} & 48.7 & 90.4\% \\
\midrule
\multicolumn{4}{l}{\emph{Dataset 3}} \\
9  & \texttt{sliding} & 46.8 & 98.1\% \\
10 & \texttt{tennis} & 54.0 & 100.0\% \\
11 & \texttt{ball-throwing-catching} & 45.4 & 100.0\% \\
\bottomrule
\end{tabular}
\end{table}

\noindent\textbf{Evaluation protocol:} For quantitative evaluation, we compare against ground-truth body poses captured by an OptiTrack motion capture system (Motive), which records 3D positions of 51 skeletal joints. To obtain mesh-based ground truth, we fit SMPL-X~\cite{SMPL-X:2019} parameters to the OptiTrack skeleton using a three-stage Adam optimization, producing per-frame ground-truth joints and mesh in the OptiTrack world frame. We report mean per-joint position error (MPJPE, cm) between predicted and ground-truth joints in OptiTrack world coordinates, and joint angle error (JAE, degrees), computed as the mean absolute angular error between predicted and ground-truth parent--child bone directions, independent of global/root pose and analogous to evaluating pose under a calibrated third-person camera setting.

\noindent\textbf{Baselines:} We compare against: \textit{(i)} SAM 3D Body~\cite{yang2026sam3dbody} using a calibrated third-person video. Compared to our system and the other egocentric baselines it leverages an additional non-wearable sensor, and thus has an unfair advantage. However, it instead suffers from occlusions and has a limited field of view; \textit{(ii)} an IMU-only baseline that uses Aria SLAM for global motion with a naive head-to-root transformation; \textit{(iii)} IMU+EgoAllo (root), which replaces only the root pose with EgoAllo while keeping the IMU-only body configuration; and \textit{(iv)} EgoAllo full egocentric body estimation. For all methods, predictions are transformed into the OptiTrack $Z$-up frame and aligned to Motive using nearest-neighbor timestamp matching in UTC. We exclude the initial calibration segment from evaluation.  
%


\subsection{Qualitative and Quantitative Evaluation}
\label{sec:results}
\begin{figure*}[t]
  \centering
  \includegraphics[width=1.9\columnwidth]{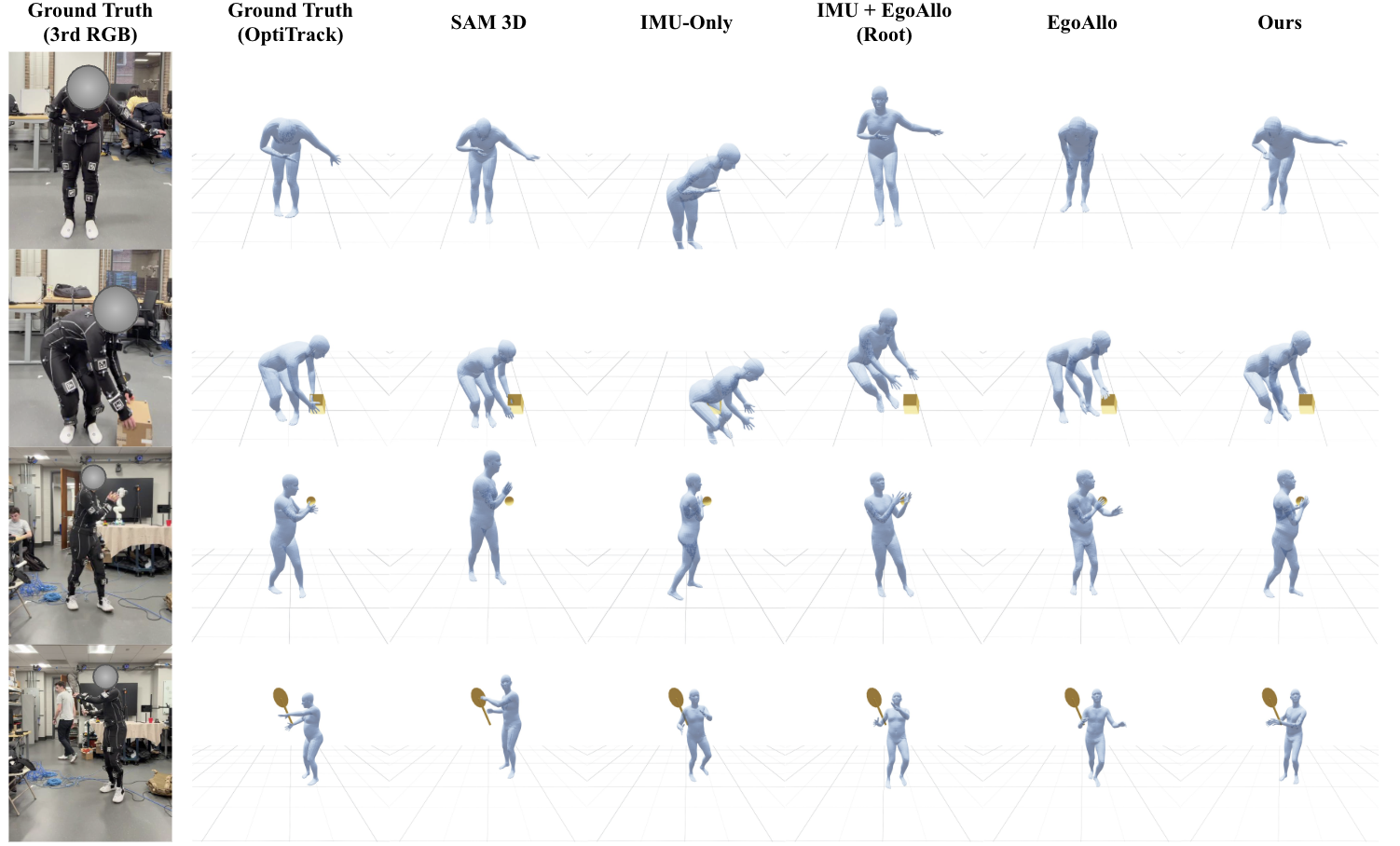}
  \caption{Qualitative 3D articulated pose results of our method and various IMU-based and third-person view-based methods, including ground truth from OptiTrack. Our method improves locomotion dynamics, foot skating and overall joint consistency compared to baselines that use IMUs or IMUs and the headset}
  \label{fig:evaluation}
\end{figure*}
\begin{table*}[t]
\centering
\caption{Quantitative results on datasets in Tab.~\ref{tab:dataset_activities} (lower is better). MPJPE is computed in the OptiTrack world frame; JAE is computed from parent--child bone directions (independent of global/root pose). In contrast to all other methods, SAM3D relies on an external calibrated camera and is therefore not a fair baseline. For SAM3D, MPJPE/JAE are computed only over frames with valid detections; its coverage is reported as recall in Tab.~\ref{tab:dataset_activities}.}
\label{tab:dataset_summary}
\small
\begin{tabular}{l|c|cccccc}
\toprule
\multirow{2}{*}{\textbf{Method}} & \multirow{2}{*}{\textbf{Egocentric}} &
\multicolumn{2}{c}{\textbf{Dataset 1}} &
\multicolumn{2}{c}{\textbf{Dataset 2}} &
\multicolumn{2}{c}{\textbf{Dataset 3}} \\
\cmidrule(lr){3-4}\cmidrule(lr){5-6}\cmidrule(lr){7-8}
&& \textbf{MPJPE (cm)} & \textbf{JAE (deg)} &
  \textbf{MPJPE (cm)} & \textbf{JAE (deg)} &
  \textbf{MPJPE (cm)} & \textbf{JAE (deg)} \\
\midrule
\textcolor{gray}{SAM3D}                 &\textcolor{red}{\ding{55}}& \textcolor{gray}{10.3} & \textcolor{gray}{10.5} & \textcolor{gray}{10.5} & \textcolor{gray}{10.7} & \textcolor{gray}{21.6} & \textcolor{gray}{11.2} \\
IMU-only (naive)      &\textcolor{green}{\ding{51}}& 16.7 & 12.6 & 18.8 & 12.2 & 16.1 & 8.9 \\
IMU + EgoAllo root    &\textcolor{green}{\ding{51}}& 12.7 & 12.5 & 11.9 & 12.2 & 12.5 & \textbf{8.7} \\
EgoAllo               &\textcolor{green}{\ding{51}}& 10.6 & 15.6 & 10.0 & 14.1 & 11.7 & 17.5 \\
\textbf{RoSHI (ours)}                &\textcolor{green}{\ding{51}}& \textbf{9.6} & \textbf{12.0} & \textbf{9.9} & \textbf{11.0} & \textbf{10.3} & 15.6 \\
\bottomrule
\end{tabular}
\end{table*}


\noindent\textbf{Qualitative:} We show qualitative results of our method in Fig.~\ref{fig:evaluation} (seventh column). We observe clear improvements in body shape reconstruction relative to EgoAllo~\cite{yi2025egoallo} (sixth column). By incorporating IMU data, our method captures locomotion dynamics more accurately and compensates for missing visual cues in cases of hand occlusion. Compared to the IMU-only baseline (fourth column), our method reduces foot skating by leveraging vision-based grounding cues. The EgoAllo root estimate further stabilizes global motion by mapping the head-mounted glasses trajectory to the pelvis with a pose-dependent head-to-pelvis transform, which adapts during leaning or squatting instead of assuming a fixed offset. Compared to IMU+EgoAllo (root, fifth column), our test-time optimization improves consistency for joints not directly observed by IMUs, yielding more coherent full-body reconstructions. We invite the reader to watch the supplementary video for additional comparisons with these baselines.

\noindent\textbf{Quantitative Results:} We show quantitative results in Tab.~\ref{tab:dataset_summary}. For completeness, we report SAM3D results in gray, but note that recall varies substantially across clips (see Tab.~\ref{tab:dataset_activities}), with the lowest coverage in Dataset~2 (e.g., 50.0\% on \texttt{pickup-walkaround}). This is largely explained by the subject moving partially or fully out of the camera field of view, so SAM3D cannot produce detections for those frames.

RoSHI achieves the best MPJPE on Datasets~1, ~2, and~3 and the best JAE on Datasets~1 and~2, showing consistent improvements in both global joint localization and articulated pose. The IMU-only baseline performs worst in terms of MPJPE across all datasets. SAM3D is competitive on frames where it successfully reconstructs the subject, but it relies on a calibrated third-person viewpoint and degrades when calibration is less accurate or when the subject is far, partially occluded, or truncated in the image.

\subsection{Humanoid Control}
\label{sec:humanoid}
\noindent\textbf{Optimization-Based Retargeting:}
As a proxy for assessing how useful our motion capture data is for humanoid control, we evaluate how easily it can be retargeted to a humanoid using standard, off-the-shelf optimization-based retargeting pipelines. In practice, RoSHI motions can be retargeted seamlessly to the Unitree G1 using existing tools such as PyRoki~\cite{pyroki2025} or GMR~\cite{araujo2025gmr}, without requiring method-specific engineering. The resulting retargeted trajectories are physically plausible, respect contact and collision constraints, and are suitable as inputs for downstream policy learning, as we demonstrate in the next section.

\begin{figure}[t]
  \centering
  \includegraphics[width=\columnwidth]{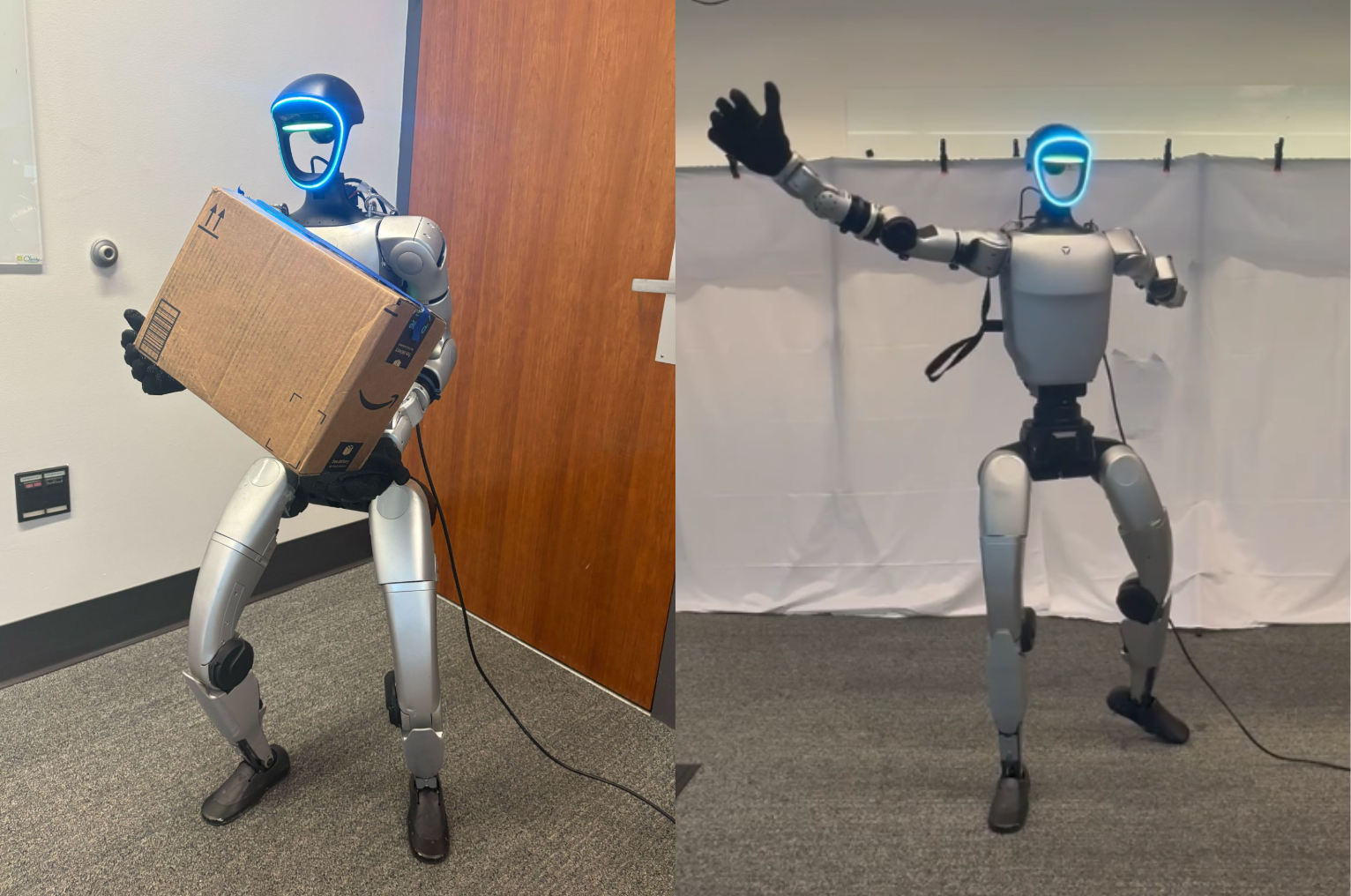}
  \caption{Deployment of the learned humanoid policy on the Unitree G1 robot. Left: robot carrying an Amazon package. Right: robot mimicking a tennis backhand racket swinging motion. We invite the reader to watch the supplementary video to see the robot in motion.}
  \label{fig:robotdeploy}
\end{figure}
\setlength{\textfloatsep}{8pt}
\setlength{\floatsep}{6pt}
\setlength{\intextsep}{6pt}

\noindent\textbf{Real-World Whole-Body Control:}
We convert re-targeted human motion sequences into reinforcement learning tracking tasks following the DeepMimic formulation \cite{Peng_2018}. Our implementation is trained in the BeyondMimic humanoid environment using its reward formulation~\cite{liao2025beyondmimicmotiontrackingversatile}, i.e., a standard tracking reward over joint angles and velocities, end-effector poses, root pose and orientation, and contact consistency. Each policy is conditioned on a global trajectory estimated by egocentric SLAM, which preserves alignment between the imitated pose and the demonstrated path and highlights that reliable localization is as important as accurate 3D body pose. With this pipeline, our wearable capture system produces data of sufficient quality to train policies that generate coordinated natural motions with accurate position control on the G1 humanoid (see Fig.~\ref{fig:robotdeploy}, and the supplementary video).


An interesting observation from the execution of highly dynamic motion policies is that the robot often tends to move at a higher velocity than the demonstrator. This is likely because the policy is conditioned on the human root trajectory, requiring the robot to reproduce not only the gait pattern but also the forward displacement. To match the trajectory of the demonstrator, it must time its gait pattern with more precision, resulting in a faster acceleration than the original demonstration. In effect, this pushes the locomotion to higher speeds and closer to the robot's physical limits.

\section{CONCLUSION}
We presented RoSHI, a portable and low-cost wearable system that fuses sparse IMUs with egocentric sensing from Project Aria to capture synchronized whole-body human motion in the wild. By combining the robustness of the inertial occlusion with global trajectory estimation based on SLAM, RoSHI prioritizes reliability and sequence stability over frame-level reconstruction accuracy, the properties most critical for humanoid policy learning. However, our system still shows improvements compared to traditional single modality only 3d body pose capture systems: the addition of IMUs improves robustness under vision occlusion, especially during agile motions. We also demonstrated that RoSHI supports diverse data collection across different whole body control sequences and that these sequences can be redirected to train reinforcement learning policies transferable to physical humanoids. Despite relying on inexpensive sensors, our data enabled successful sim-to-real transfer of dynamic behaviors such as running and jumping, as well as expressive gestures including bowing and waving.\\
\\
\noindent\textbf{Limitations:} Since IMUs cannot directly observe all joint degrees of freedom, the reconstruction quality for unobservable joints may degrade, particularly under ambiguous motion. Moreover, when multiple constraints with conflicting directional deviations are imposed, the optimization may introduce twisting artifacts, forcing less-constrained joints into unnatural configurations. In such cases, the most affected joints are those along the kinematic chain between the shoulder and the pelvis, as they are not directly observable from the IMU measurements. We believe that addressing these limits can be the subject of future work. 

\section{ACKNOWLEDGMENTS}
This work was supported by DARPA under Agreement No. HR0011-24-9-0430, and the Swiss National Science Foundation under Grant No. 225354. We thank the Meta Aria team for their support and for providing access to the Aria hardware and software infrastructure.






\bibliographystyle{IEEEtran}
\bibliography{reference} 

\end{document}